%% file: sample_paper.tex
\documentclass[twoside]{article}

\usepackage[accepted]{aistats2020}

\input{command-file-arXiv}

\usepackage{algorithm}
\usepackage{algorithmic}
\usepackage{subfloat}
\usepackage{subfig}

\usepackage{pgfplots}
\pgfplotsset{width=5.45cm,compat=1.14}

\begin{document}

%
\runningtitle{AM converges super-linearly for mixed regression}

%
\runningauthor{Ghosh and Ramchandran}

\twocolumn[

\aistatstitle{Alternating Minimization Converges Super-Linearly for  Mixed Linear Regression}
\vspace{-5mm}
\aistatsauthor{ Avishek Ghosh$^*$ \And Kannan Ramchandran$^*$}
\vspace{3mm}
\aistatsaddress{$^*$Department of Electrical Engineering and Computer Sciences, UC Berkeley \\
avishek$\_$ghosh@berkeley.edu, kannanr@eecs.berkeley.edu} 
]

\vspace{-3 mm}
\begin{abstract}
  We address the problem of solving mixed random linear equations. We have unlabeled observations coming from multiple linear regressions, and each observation corresponds to exactly one of the regression models. The goal is to learn the linear regressors from the observations. Classically, Alternating Minimization (AM) (which is a variant of Expectation Maximization (EM)) is used to solve this problem. AM iteratively alternates between the estimation of labels and solving the regression problems with the estimated labels. Empirically, it is observed that, for a large variety of non-convex problems including mixed linear regression, AM converges at a much faster rate compared to gradient based algorithms. However, the existing theory suggests similar rate of convergence for AM and gradient based methods, failing to capture this empirical behavior. In this paper, we close this gap between theory and practice for the special case of a mixture of $2$ linear regressions.  We show that, provided initialized properly, AM enjoys a \emph{super-linear} rate of convergence in certain parameter regimes. To the best of our knowledge, this is the first work that theoretically establishes such rate for AM. Hence, if we want to recover the unknown regressors upto an error (in $\ell_2$ norm) of $\epsilon$, AM only takes $\mathcal{O}(\log \log (1/\epsilon))$ iterations. Furthermore, we compare AM with a gradient based heuristic algorithm empirically and show that AM dominates in iteration complexity as well as wall-clock time.
\end{abstract}

\section{INTRODUCTION}
\label{sec:intro}

We assume that the measurements are coming from the following observation model:
 \begin{align}
 y_i = \obs{i}{1} z_i + \obs{i}{2} (1-z_i) + w_i,\label{eqn:model}
 \end{align}
 for $i = 1,\ldots, n$, where the covariates are $\{\meas_i\}_{i=1}^n \in \real^d$ and the unknown regressors are $\sigstar_1$ and $\sigstar_2$. $z_i$ here is the (unknown) latent variable taking values $\{0,1\}$. When $z_i = 1$, the $i$-th observation comes from the regressor $\sigstar_1$, and $z_i = 0$ implies $y_i$ coming from $\sigstar_2$. $w_i$ here denotes the additive noise. Given the covariate-response pairs $\left( \meas_i,y_i \right)_{i=1}^n$, the goal is to estimate $(\sigstar_1,\sigstar_2)$ without the knowledge of $\{z_i\}_{i=1}^n$.

Let us provide some motivation for studying the model ~ \eqref{eqn:model}. When measurements are obtained from multiple latent classes and the goal is to estimate the underlying parameters, mixed linear regression is often a reasonable model to assume. The model is introduced by Wedel et al (1995) (\cite{wedel1995mixture}) and have become a standard framework for applications like health-care \cite{health}, market segmentation \cite{wedel2012market} and music perception \cite{viele2002modeling}. Please refer to 
Grun et al (2007) (\cite{grun2007applications}) and the references therein for several other applications of the mixture model. Mixed regression model also serves as a theoretical tool for analyzing benchmark nonconvex optimization algorithms (\cite{chaganty2013spectral}; \cite{Klusowski}) or analyzing new algorithms (\cite{chen_convex}). Furthermore, the mixed regression model is a close relative of the classical hierarchical mixtures of experts \cite{jacobs1991adaptive}, which has several applications in Generative Adversarial Networks (GAN) and Gated Recurrent Units (GRU) (\cite{ashok}).

A generic way to approach problems with latent variables is via the EM algorithm or its variants. Alternating Minimization (AM), which can be thought as \emph{hard-EM} is classically used to solve  \eqref{eqn:model}. In every iteration AM first guesses the labels, and subsequently solve the linear regression problems with the guesses labels. With Gaussian covariates and a proper initialization, AM provably converges to the optimal parameters at a linear rate. In Yi et al (2014) (\cite{yitwo}), AM for mixed linear regression was proposed and analyzed for the problem of $2$ mixtures (similar to \eqref{eqn:model}), and later it has been extended to the setting with more than $2$ mixtures (\cite{yimany}). A few recent works (\cite{high_dim_em}, \cite{zhu2017high}, \cite{improved_chen}, \cite{chen2018covariate} \cite{Klusowski}, \cite{caramanis_em} ) also use AM (or its variant EM) to tackle different aspects of the mixed linear regression and related problems.

Another line of work uses gradient descent algorithm to solve \eqref{eqn:model}.  Although solving a global non-convex problem (owing to unknown labels) Zhong et al (2016) (\cite{jain_gd}) shows that under certain assumptions, the problem is locally strongly convex and hence gradient descent converges at a linear rate. Later Li et al (2018) (\cite{li_gd}) improves the sample and computational complexity via a careful analysis of the gradient descent algorithm. Furthermore, Chaganty et al (2013) (\cite{chaganty2013spectral}) and  Sedghi et al (2016) (\cite{sedghi2016provable}) use tensor method to solve the mixed linear problem and Chen et al (2014) (\cite{chen_convex}) provides a convex relaxation formulation of a mixture of $2$ regression problem and proposes a mini-max optimal algorithm. However, the tensor decomposition based method or the nuclear norm based convex relaxation method is computation heavy and slow.

Alternating Minimization is a general purpose algorithm used to solve non-convex problems with latent variables. A few examples of such problems include phase retrieval \cite{netrapalli2013phase}, matrix sensing and completion \cite{jain_low} and max-affine regression \cite{ghosh_ma}. With proper initialization, it is empirically observed that AM is much faster than gradient based algorithms. For example, in the context of phase retrieval problem, Table 1 of Zhang et al (2016) (\cite{zhang2016reshaped}) shows that gradient descent takes $36 \times$ more iterations and $2.36 \times$ more wall-clock time compared to AM. Using a truncation and reshaping technique particular to the phase retrieval problem, Zhang et al (2016) reduces the wall-clock time but still requires $12 \times$ iterations over AM. It was also conjectured in Xu et al (1996) (\cite{xu1996convergence}), that EM (i.e., soft-AM) enjoys a super-linear rate of convergence, much like the Newton method for convex optimization. Also, Balakrishnan et al (2017) (\cite{balakrishnan2017statistical}) observes a very fast rate of convergence of EM when initialized properly for the problem of estimation from mixture of Gaussians. Later Daskalakis et al (2017) (\cite{daskalakis2017ten}) shows that it is sufficient to run EM for only constant number of iterations for the Gaussian mixture problem, thus hinting towards a meteoric speed of convergence.  However, to the best of our knowledge, there is no theoretical justification to this fact for either EM or AM.

The goal of this paper is to bridge this gap between theory and practice. We consider the classical AM algorithm to solve the mixed linear regression with $2$ components. Via a careful analysis of the underlying empirical process of the AM iteration, we prove that the rate of convergence of AM in this setting is truly \emph{super-linear}, thus explaining the empirical phenomenon mentioned above. We believe that this is the first work to theoretically establish such faster rate of convergence. We also run exhaustive experiments to show the super-linear rate of AM, and compare with a gradient based heuristic. We observe that AM dominates the gradient based heuristic in terms of both iteration complexity and wall-clock time.

Very recently, Shen et al (2019) (\cite{shen2019iterative}) uses an iterative least trimmed squares (ILTS) algorithm for the mixture of regressions. Based on residual values, ILTS iteratively estimates the components of mixture one-by-one and removes the corresponding observations in subsequent iterations. Under certain settings, ILTS is provably shown to converge super-linearly. Note that, algorithmically, the classical AM is very different from ILTS. There is no (direct) latent variable estimation phase in ILTS. Also AM estimates the all regressions at once, instead of iterative estimation and trimming. Furthermore, the statistical tools and techniques we use are quite different from \cite{shen2019iterative}.

\paragraph{Our Contributions:} We have the following:
\begin{itemize}
\item We analyze the classical AM (Algorithm~\ref{alg:am}) for a mixture of $2$ regressions, and show that provided initialized properly, the rate of convergence is \emph{super-liner} with exponent $3/2$ (Theorem 1). Instead of using crude singular value bounds, we fine-tune the underlying empirical process and obtain a better convergence rate.  The sample complexity required for our algorithm is $n \geq C d$ (where $C$ is a universal constant), which is information theoretically optimal (\cite{yitwo}).

\item Upon further polishing the analysis of AM iterates, we identify a regime where the rate of convergence is quadratic (with exponent $2$), which is identical to Newton method for convex optimization (Theorem 3).

\item Via numerical experiments (Section~\ref{sec:simulations}), we demonstrate the super-linear convergence of AM. Although we consider a mixture of $2$ regression in theory, we show in simulations that more than $2$ mixtures of linear regression also enjoys the super linear convergence. Also, we compare the performance of AM with a gradient based heuristic (Algorithm~\ref{alg:heu}). We show that the rate of convergence of the gradient based heuristic is linear. Furthermore, we observe that on average the gradient based heuristic takes $8 \times$ iterations and $6 \times$ wall-clock time compared to AM.
\end{itemize}

\subsection{Notations}
We use $\| .\|$ to denote the $\ell_2$ norm of a vector unless otherwise specified. $[n]$ denotes the set of integers $\{1,2,\ldots, n\}$. Throughout the paper, we use $C,C_1,C_2,..,c,c_1,c_2 ..$ to represent positive universal constants, the value of which may change from instance to instance.  

\section{AM FOR THE MIXTURE OF REGRESSIONS}
\label{sec:algo}

In this section, we describe the AM algorithm for parameter estimation where the observations are coming from \eqref{eqn:model}. In particular, we are interested in solving the following least squares problem:
\begin{align}
&L(\theta_1,\theta_2) = \sum_{i=1}^n \min_{z_i \in \{0,1\}}  \bigg (y_i - \langle \meas_i, z_i \theta_1 + (1-z_i) \theta_2 \rangle \bigg )^2, \nonumber \\
&(\widehat{\theta}^{ls}_1,\widehat{\theta}^{ls}_2) = \mathrm{argmin}_{\theta_1 \in \real^d,\theta_2 \in \real ^d} \,\, L(\theta_1,\theta_2) \label{eqn:least_squares}
\end{align}
where $\{\meas_i\}_{i=1}^n$ are the covariates and $\{z_i\}_{i=1}^n$ are the latent variables. Note that the above problem is non-convex owing to the presence of $\{z_i\}_{i=1}^n$. Furthermore, \eqref{eqn:least_squares} is NP-hard for general covariates $\{x_i\}_{i=1}^n$ (\cite{yitwo}). However the problem becomes tractable with structured covariates (e.g., i.i.d Gaussian covariate) and proper initialization. We use Alternating Minimization (AM) to solve the least squares problem of \eqref{eqn:least_squares}, the steps of which are described in Algorithm~\ref{alg:am} (also Algorithm 1 of \cite{yitwo}).

\begin{algorithm}[t]
  \caption{AM for mixed linear regression}
  \begin{algorithmic}[1]
  \INPUT Covariate response pairs $(\meas_i,y_i)_{i=1}^n$, initialization $(\widehat{\theta}^{(0)}_1,\widehat{\theta}^{(0)}_2)$, number of rounds $T$

  \STATE Split samples in $T$ disjoint groups $(x_i^{(t)},y_i^{(t)})_{t=1}^{n/T}$, where  $t = 0,1 ,\ldots,T-1$
  
   \STATE \textbf{for} $t = 0,\ldots, T-1$ \textbf{do} 
   \STATE \hspace{ 4mm} $z_i^{(t)} = \mathrm{argmin}_{j \in \{0,1\}} | y_i^{(t)} - \langle x_i^{(t)}, \widehat{\theta}^{(t)}_j \rangle|;\, i \in [n/T]$
   \STATE \hspace{4mm} $\widehat{\theta}^{(t+1)}_j = \mathrm{argmin}_{\theta} (y_i^{(t)} - \langle x_i^{(t)},\theta \rangle )^2 \ind{z_i^{(t)} = j};$ \\
   \hspace{20 mm} for $j \in \{0,1\}$
   
   \STATE \textbf{end for}
 
  \OUTPUT $(\widehat{\theta}^{(T)}_1,\widehat{\theta}^{(T)}_2)$
  \end{algorithmic}\label{alg:am}
\end{algorithm}

The first step of Algorithm~\ref{alg:am} is sample-splitting across iterations. The sample split step is standard in the theoretical analysis of AM. For example, \cite{yitwo} and \cite{yimany} use sample split for mixture of regressions, \cite{netrapalli2013phase} uses it for phase retrieval and \cite{jain_low} uses it for matrix completion. As illustrated in Section~\ref{sec:simulations}, we do not require sample-split in experiments. This assumption is only for theoretical tractability. Also, since the AM converges in super-linear speed, the sample-split will only increase the sample complexity, $n$, by a multiplicative factor of $\log \log (1/\epsilon)$, where $\epsilon$ is the tolerable error (in $\ell_2$ norm) in the recovery of $(\theta^*_1,\theta^*_2)$. In the above-mentioned problems, sample split results in the increase of sample complexity by a multiplicative factor of $\log (1/\epsilon)$, which is much larger compared to the price we pay.

As seen in Algorithm~\ref{alg:am}, each iteration of AM consists of $2$ steps. First, the labels $\{z_i\}_{i=1}^n$ are estimated. This is done by calculating which regressor estimate yields the linear model closer to the observation. Once the label ambiguity is resolved, the problem is now converted to $2$ ordinary least squares. The solutions of the least squares yield the next iterate.

\section{MAIN RESULTS}
\label{sec:main_results}

We now present the main results of the paper. We characterize the convergence rate of Algorithm~\ref{alg:am}. We make the following structural assumption on the covariates, which is standard and featured in several previous works (\cite{yitwo,yimany,netrapalli2013phase,ghosh_ma}).

\begin{assumption}
The covariates $\{\meas_i\}_{i=1}^n$ are drawn i.i.d from the standard $d$-dimensional Gaussian distribution, $(\meas_i \stackrel{i.i.d}{\sim} \NORMAL(0,I_d))$.
\end{assumption}

 In this section, we also assume $w_i = 0$ for all $i \in [n]$. We emphasize here that in simulations (Section~\ref{sec:simulations}), we observe that our theory perfectly holds even in the presence of noise. However, for analysis we deal with the noiseless scenario only. 
 
 Let $p_1$ and $p_2$ be the fraction of observations coming from $\sigstar_1$ and $\sigstar_2$ respectively. We also define the following error metric to quantify the closeness of AM iterates at $t$-th iteration to the true parameters $(\sigstar_1,\sigstar_2)$: 
 \begin{align*}
 \mathsf{dist}(\sigit{t}_1,\sigit{t}_2):= \max \{ \|\sigit{t}_1 - \sigstar_1 \|, \|\sigit{t}_2 - \sigstar_2 \| \}.
 \end{align*}
To simplify notation, we drop the superscript from $\sigit{t}_i$ for $i=1,2$ and consider one iteration of the AM algorithm with $(\theta_1,\theta_2)$ as input and $(\theta_1^+,\theta_2^+)$ as output. It is sufficient to show the one step contraction for Algorithm~\ref{alg:am}. Furthermore, let $n/T:=n_1$, where $n$ and $T$ are the sample complexity and the number of iterations of Algorithm~\ref{alg:am} respectively. Hence, the number of samples for this particular iteration is $n_1$. We have the following result.

\begin{theorem}
\label{thm:sup_lin}
Suppose $n_1 \geq C d$, and the following 
\begin{align*}
\mathsf{dist}(\theta_1,\theta_2) \leq \frac{\|\theta^*_1 - \theta^*_2 \|}{2 \log n_1}
\end{align*}
holds. Then, the inequality
\begin{align*}
\mathsf{dist}^2(\theta_1^+,\theta_2^+) \leq  \left[ \frac{\log n_1}{4 \|\theta^*_1-\theta^*_2 \|}\right ] \mathsf{dist}(\theta_1,\theta_2)^{3}
\end{align*}
is satisfied with probability exceeding $1-c_1 n_1^{-10}$ provided $\mathsf{dist}(\theta_1,\theta_2) \geq c_2 \max\{p_1,p_2\} \sqrt{\frac{\log n_1}{n_1}}$.
 \end{theorem}
\begin{figure}[t!]
    \centering
    \includegraphics[height = 0.11\textwidth, width=0.4\textwidth]{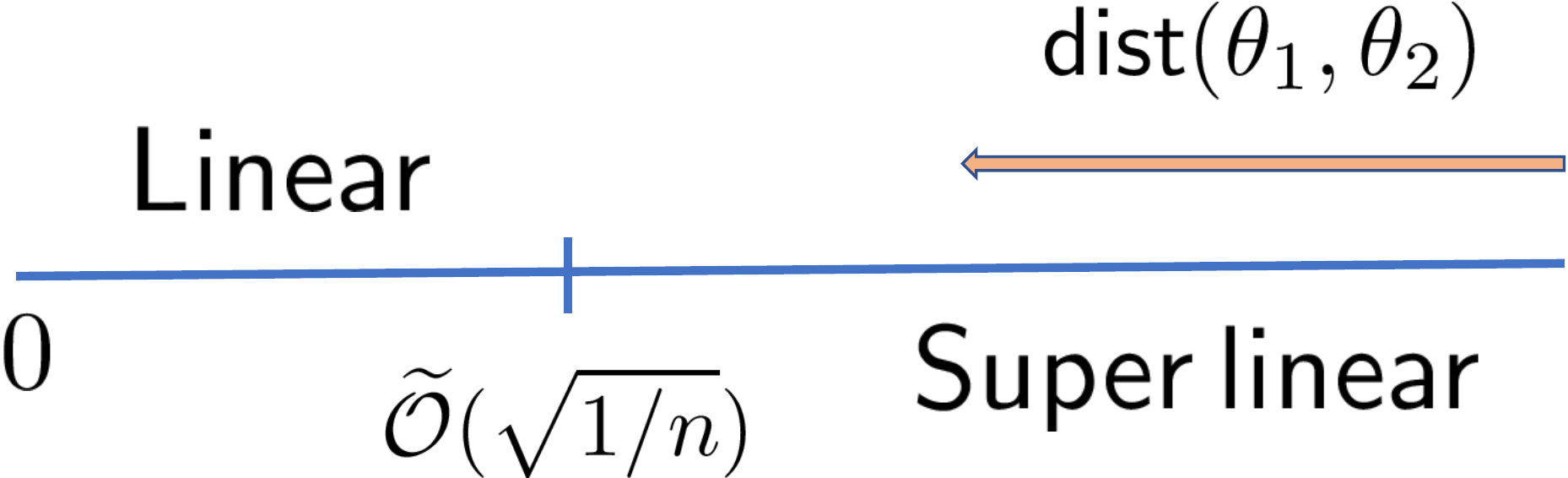}
    
    \caption{Convergence region for iterates of AM algorithm. Super-linear region corresponds to convergence with exponent $3/2$.
     }
    \label{fig:illust_one}
   
\end{figure}

The above theorem shows the super-linear rate of convergence of the $\mathsf{dist}$ function. Hence, for an error tolerance (in $\ell_2$ norm) of $\epsilon$, we have $T = \mathcal{O}(\log \log (1/\epsilon))$. Note that the super-linear rate holds as long as $\mathsf{dist}(\theta_1,\theta_2) \geq \widetilde{\mathcal{O}}(\sqrt{1/n})$. If $\mathsf{dist}(\theta_1,\theta_2)$ falls below the mentioned threshold, the rate of convergence is no longer super-linear; it falls back to the linear regime (Figure~\ref{fig:illust_one}). This is because the concentration inequalities we use to prove Theorem~\ref{thm:sup_lin} cease to produce any meaningful results if $\mathsf{dist}(\theta_1,\theta_2) < \widetilde{\mathcal{O}}(\sqrt{1/n})$.

The proof of Theorem~\ref{thm:sup_lin} is deferred to Section~\ref{sec:proof_thm_one}. Super linear convergence is a consequence of two facts: 
\begin{enumerate}
\item The fraction of non-zero (or \emph{active}) terms that contribute in $\mathsf{dist}(\theta_1^+,\theta_2^+)$ is bounded by $\mathcal{O}(\mathsf{dist}^{1/2}(\theta_1,\theta_2))$.
\item Each active term is $\mathcal{O}(\mathsf{dist}(\theta_1,\theta_2))$.
\end{enumerate}

In the prior works (e.g., \cite{yitwo}), using crude singular value based technique, the fraction of active terms were bounded by $\mathcal{O}(1)$. Since each active terms contribute $\mathcal{O}(\mathsf{dist}(\theta_1,\theta_2))$, a linear rate is obtained. We instead use (sharp) empirical process tools to capture the \emph{active} terms and improve the rate of convergence.

If the desired error tolerance (in $\ell_2$ norm) is much less than $\widetilde{\mathcal{O}}(\sqrt{1/n})$, then Theorem~\ref{thm:sup_lin} fails to characterize the entire behavior of the iterates. In order to so, we  appeal to \cite[Theorem 1]{yitwo} which yields the following linear rate. 
\begin{theorem}
\label{thm:lin}
 Suppose that $n_1 \geq (C/\min\{p_1,p_2\}) d$. Then, provided $\mathsf{dist}(\theta_1,\theta_2) \leq c \min\{p_1,p_2\} \|\theta^*_1 - \theta^*_2 \|$,
the inequality
$\mathsf{dist}(\theta_1^+,\theta_2^+) \leq \frac{1}{2} \mathsf{dist}(\theta_1,\theta_2)$
holds with probability greater than $1-c \exp\{-c_1 d\}$.
\end{theorem}

\subsection{Faster Rates: Quadratic Rate of Convergence}

We now prove an improved rate of convergence for AM. In particular, we show that in a particular regime, the convergence rate of AM is quadratic (with exponent $2$), which is an improvement over the rate in Theorem~\ref{thm:sup_lin}. We have the following result.
\begin{theorem}
\label{thm:quadratic}
Suppose that $n_1 \geq (C/\min\{p_1,p_2\}) d$, and the following
\begin{align*}
\mathsf{dist}(\theta_1,\theta_2) \leq c \min\{p_1,p_2 \} \|\theta^*_1 - \theta^*_2 \|
\end{align*} 
holds. Then the inequality
\begin{align*}
\mathsf{dist}(\theta_1^+,\theta_2^+) \leq \frac{1}{2} \mathsf{dist}(\theta_1,\theta_2)^{2}
\end{align*}
holds with probability exceeding $1-c_1 \exp\{-c_2 d \} - c_3 n_1^{-10}$ provided \\
 $\mathsf{dist} \geq \max \left \lbrace C \max\{p_1,p_2\} \sqrt{\frac{\log n_1}{n_1}}, \frac{d}{C_1 n_1}\right \rbrace $.
\end{theorem}
 
The proof is deferred to the Supplementary material. The gain in rate comes from a careful analysis of the spectrum of a Gaussian random matrix in conjuction with the fact that the fraction of \emph{active} terms is $\mathcal{O}(\mathsf{dist}^{1/2}(\theta_1,\theta_2))$.

\begin{figure}[t!]
    \centering
    \includegraphics[height = 0.13\textwidth, width=0.43\textwidth]{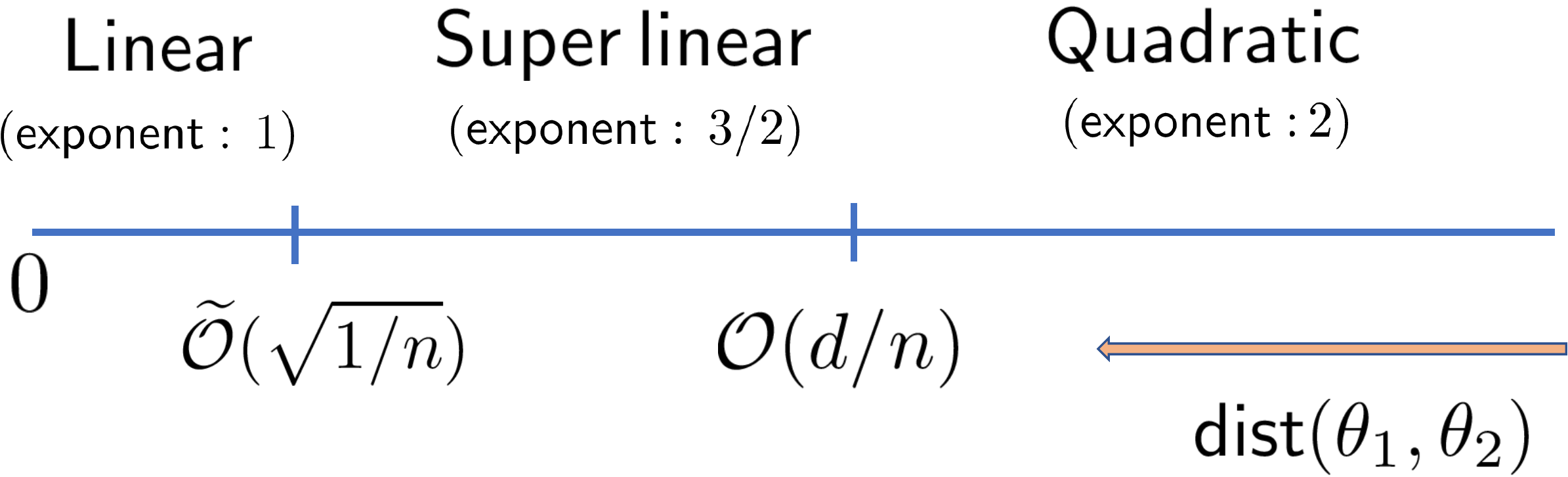}
    
    \caption{Convergence region for iterates of AM algorithm. Quadratic, Super-linear and linear region of convergence is shown.
     }
    \label{fig:illust_two}
   
\end{figure}

 Combining Theorem~\ref{thm:sup_lin} and \ref{thm:quadratic}, we are now able to completely characterize the convergence behavior of the iterates of the AM algorithm. It is shown in Figure~\ref{fig:illust_two}. Until $\mathcal{O}(d/n)$, the convergence is quadratic. Beyond this point the rate slows down but maintains a super-linear rate upto $\widetilde{\mathcal{O}}(\sqrt{1/n})$. After this point, the convergence speed slows down to a linear rate.

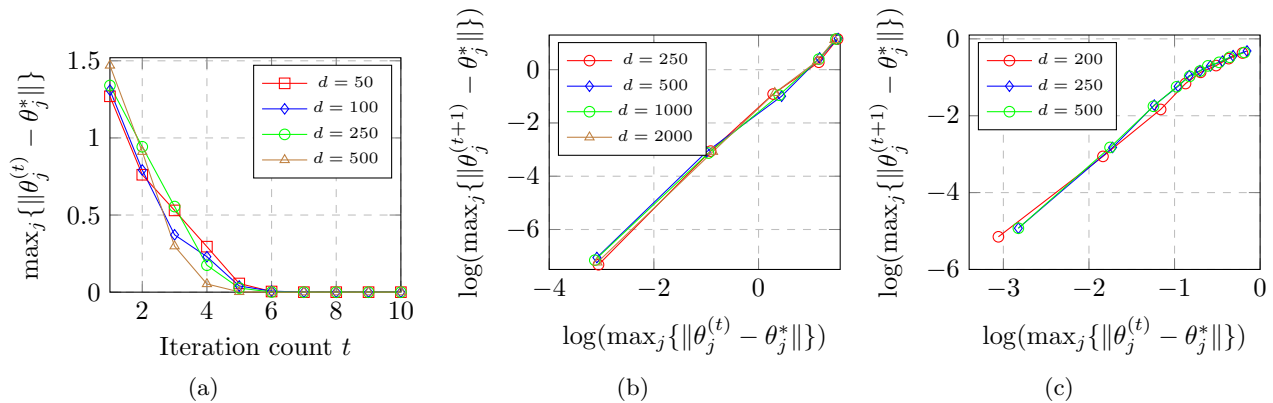
\begin{figure*}[t!]
\centering
\subfigure[]{
\begin{tikzpicture}
\begin{axis}[legend style={font=\tiny},
 xlabel={Iteration count $t$},
    ylabel={$\max_j \{\| \theta^{(t)}_j - \theta^*_j \| \}$},
    xmin=1, xmax=10,
    ymin=0, ymax=1.52,
    legend pos= north east,
    ymajorgrids=true,
    xmajorgrids=true,
    grid style=dashed,
]
 
\addplot[
    color=red,
    mark=square,
    ]
    coordinates {
    (1,1.27)(2,7.61245328e-01 )(3,5.30075041e-01)(4,2.94999854e-01)(5,5.62291869e-02)(6,4.32321318e-03)(7,2.84490181e-05)(8,0
)(9,0)(10,0)
    };
    \addlegendentry{$d =50$}

\addplot[
    color=blue,
    mark=diamond,
    ]
    coordinates {
    (1,1.31)(2,7.92135992e-01 )(3, 3.71716758e-01)(4,2.29829755e-01)(5,4.00340620e-02)(6,6.86442120e-03)(7,0)(8,0)(9,0)(10,0)
    };
    \addlegendentry{ $d = 100$}
    
  \addplot[
    color=green,
    mark=o,
    ]
    coordinates {
    (1,1.34 )(2, 9.42748583e-01 )(3,5.54962613e-01)(4,1.74634098e-01)(5, 2.66790665e-02 )(6,9.98894986e-04)(7,0)(8,0)(9, 0)(10,0)
    };
    \addlegendentry{$ d =250 $}

    \addplot[
    color= brown,
    mark=triangle,
    ]
    coordinates {
    
    (1,1.47)(2,9.10579176e-01)(3, 2.98125177e-01 )(4, 5.30779715e-02 )(5, 1.94205066e-03 )(6, 5.69070109e-14)(7,  0)(8, 0)(9,0)(10, 0)
    };
    \addlegendentry{$d = 500$}
    
\end{axis}
\end{tikzpicture}}
\subfigure[]{
\begin{tikzpicture}
\begin{axis}[legend style={font=\tiny},
    xlabel={$\log ( \max_j \{ \|\theta^{(t)}_j - \theta^*_j \| \} )$},
    ylabel={$\log ( \max_j \{ \|\theta^{(t+1)}_j - \theta^*_j \| \})$ },
    xmin=-4, xmax=1.56,
    ymin=-7.5, ymax=1.3,
    legend pos=north west,
    ymajorgrids=true,
    xmajorgrids=true,
    grid style=dashed,
]
 
\addplot[
    color=red,
    mark=o,
    ]
    coordinates {
  (1.51, 1.1512)  (1.15129255 ,0.27850234)  (0.27850234, -0.9156381)( -0.9156381 ,-3.05812329) ( -3.05812329, -7.3233779) 
    };
    \addlegendentry{$d=250$}

\addplot[
    color= blue,
    mark=diamond,
    ]
    coordinates {
    (1.52, 1.16788) (1.16786614, 0.44108709) ( 0.44108709,-0.99662787) (-0.996662787,-3.08899077) ( -3.08899077,-7.05261227)
    };
    \addlegendentry{$d=500$}

    \addplot[
    color=green,
    mark=o,
    ]
    coordinates {
   (1.48, 1.15) (1.151, 0.394108709) ( 0.39410870,-0.9566) (-0.956662787,-3.12899077) ( -3.12899077,-7.15261227)
    };
    \addlegendentry{$d=1000$}
    
    \addplot[
    color=brown,
    mark=triangle,
    ]
    coordinates {
   (1.45,1.12) (1.12 ,0.31850234)  (0.31, -0.87156381)( -0.87156381 ,-3.07812329) ( -3.07812329, -7.23233779)
    };
    \addlegendentry{$d=2000$}

\end{axis}
\end{tikzpicture}}
\subfigure[]{
\begin{tikzpicture}
\begin{axis}[legend style={font=\tiny},
    xlabel={$\log ( \max_j \{ \|\theta^{(t)}_j - \theta^*_j \| \} )$},
    ylabel={$\log ( \max_j \{ \|\theta^{(t+1)}_j - \theta^*_j \| \} )$ },
    xmin=-3.4, xmax=0,
    ymin=-6, ymax=0.1,
    legend pos=north west,
    ymajorgrids=true,
    xmajorgrids=true,
    grid style=dashed,
]
 
\addplot[
    color=red,
    mark=o,
    ]
    coordinates {
  (-0.20611706, -0.36785874)(-0.36785874, -0.51470194)(-0.51470194, -0.69520516)(-0.69520516, -0.87153345)(-0.87153345, -1.16260697)(-1.16260697, -1.8363821)( -1.8363821, -3.0580655)(-3.0580655 , -5.15013962)

    };
    \addlegendentry{$d=200$}

\addplot[
    color= blue,
    mark=diamond,
    ]
    coordinates {
    ( -0.15314174,-0.31555399) ( -0.31555399,  -0.43949868) ( -0.43949868,-0.58079301) (-0.58079301,-0.70286525)( -0.70286525,-0.82844448)( -0.82844448,-0.95837261)( -0.95837261, -1.23653953)( -1.23653953,-1.72357844)(
       -1.72357844,-2.82483505)( -2.82483505,-4.92643918)
    };
    \addlegendentry{$d=250$}

    \addplot[
    color=green,
    mark=o,
    ]
    coordinates {
   ( -0.18314174,-0.35555399) ( -0.35555399,  -0.47949868) ( -0.47949868,-0.61079301) (-0.61079301,-0.70286525)( -0.70286525,-0.83844448)( -0.8234448,-0.975837261)( -0.975837261, -1.24653953)( -1.24653953,-1.75357844)(
       -1.75357844,-2.82483505)( -2.82483505,-4.92643918)
    };
    \addlegendentry{$d=500$}
    \end{axis}
\end{tikzpicture}}
\caption{Convergence of the AM with Gaussian covariates---in panel (a), for a mixture of $2$ linear regression, we plot the distance to the true parameters $\max_{j \in 1,2} \{ \|\theta^{(t)}_j - \theta^*_1 \| \}$ over iterations $t$ for different $d$ ($50,100,250,500$), where we set $n =6 \,d$. Panel (b) shows the super-linear convergence of AM (with exponent $1.7 - 1.8$) for mixed regression with $2$ components. Here we choose $d = 250,500,1000,2000$ and $n = 6 \, d$. In Panel (c), we consider the problem of mixed linear regression with $3$ components. Keeping $n = 15\, d$ and varying $d (200,250,500)$, we show that AM retains the super-linear convergence for more than $2$ mixtures of linear regression. All the points in the plots are obtained via taking average over $20$ trials.}
\label{fig:am_conv}
\end{figure*}

\section{INITIALIZATION}
\label{sec:ini}
As seen in Theorem~\ref{thm:sup_lin},\ref{thm:lin} and \ref{thm:quadratic}, the convergence guarantee of AM requires the initial values of the iterate to be close to the optimal parameters. In particular, we need the initial values to be within a norm ball of constant radius of the optimal parameters.

Usually, spectral methods are employed for initializing AM for a large class of problems. Since we have the covariate-response pairs $(\meas_i, y_i)_{i=1}^n$, we can compute an appropriate matrix, and the singular value decomposition (SVD) of the matrix yields required initialization. In \cite{candes2013phaselift}, \cite{netrapalli2013phase}, \cite{walds}, the spectral method of initialization is used for the phase retrieval problem, and in \cite{ghosh_ma}, it is used for the max-affine regression problem.

In Algorithm 2 of Yi et al (2014) (\cite{yitwo}), a spectral method for the mixture of $2$-component mixture is provided. The algorithm first constructs a matrix $M = \sum_{i=1}^n y_i^2 \meas_i \meas_i^\top$. Then, using SVD, the subspace spanned by the eigenvectors corresponding to the top $2$ eigenvalues is obtained. It is shown in \cite{yitwo} that the optimal parameters $(\theta^*_1,\theta^*_2)$ lie in this subspace. Finally, \emph{griding} the $2$-dimensional subspace yields in the required initial values of the iterates of AM.

For the mixture of regression with more than $2$ components, Yi et al (2016) (\cite{yimany}) uses a  tensor decomposition technique. After obtaining the subspace from appropriate matrix, a tensor is constructed from the covariate-response pairs $(\meas_i, y_i)_{i=1}^n$. Decomposing the tensor results in the required initialization.

In this paper, we use a slightly stricter initialization (by a log factor) of \cite{yitwo} to get the conditions of Theorem~\ref{thm:sup_lin},\ref{thm:lin} and \ref{thm:quadratic}. When $n/(\log n)^2 \sim d(\log d)^2$, we get this initialization by using \cite[Proposition 2]{yitwo}. Since we work with $2$ components, employing a tensor decomposition method of \cite{yimany} is unnecessary because griding a $2$-dimensional subspace requires very light computation.

\section{SIMULATIONS}
\label{sec:simulations}
In this section, we validate the theory presented in Section~\ref{sec:main_results}. Additionally, we handle the setting where the observations, $\{y_i\}_{i=1}^n$ are corrupted with additive noise. Furthermore, we consider the problem of mixed linear regression with more than $2$ components. Note that as mentioned in Section~\ref{sec:main_results}, in all our experiments, we do not require the sample-split step of Algorithm~\ref{alg:am}. Finally, we compare the performance of AM with a gradient based heuristic.

For the following experiments, we sample the covariates $\{\meas_i \}_{i=1}^n$ in an i.i.d fashion from the standard $d$-dimensional Gaussian distribution. We randomly select the $d$-dimensional true parameters $\{\theta^*_1,\theta^*_2\}$. The observations are then obtained via equation~\eqref{eqn:model}.

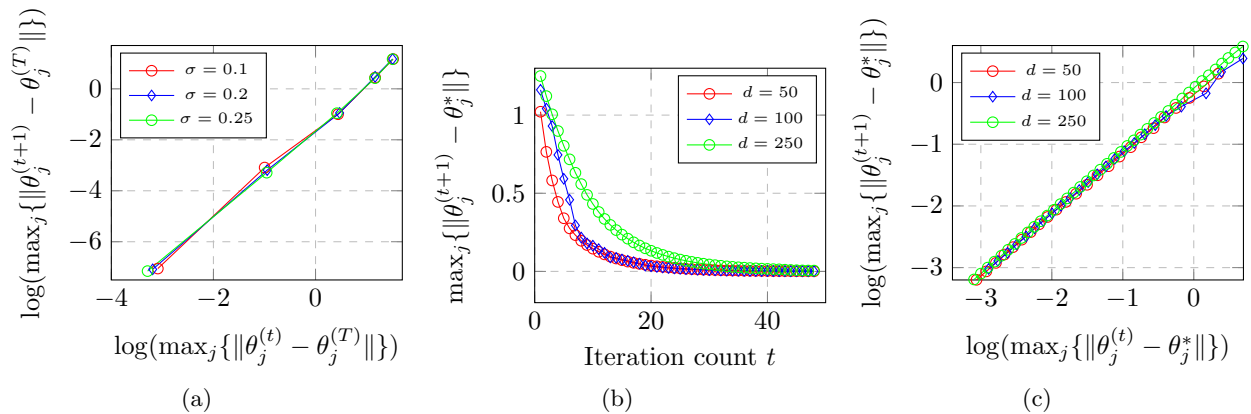
\begin{figure*}[t!]
\centering
\subfigure[]{
\begin{tikzpicture}
\begin{axis}[legend style={font=\tiny},
    xlabel={$\log ( \max_j \{ \|\theta^{(t)}_j - \theta^{(T)}_j \| \} )$},
    ylabel={$\log ( \max_j \{ \|\theta^{(t+1)}_j - \theta^{(T)}_j \| \})$ },
    xmin=-4, xmax=1.7,
    ymin=-7.5, ymax=1.7,
    legend pos=north west,
    ymajorgrids=true,
    xmajorgrids=true,
    grid style=dashed,
]
 
\addplot[
    color=red,
    mark=o,
    ]
    coordinates {
    (1.52, 1.16788) (1.16786614, 0.44108709) ( 0.44108709,-0.99662787) (-0.996662787,-3.08899077) ( -3.08899077,-7.05261227)
    };
    \addlegendentry{$\sigma = 0.1$}

\addplot[
    color= blue,
    mark=diamond,
    ]
    coordinates {
    (1.51, 1.16788) (1.1671786614, 0.47108709) ( 0.47108709,-0.94662787) (-0.946662787,-3.1899077) ( -3.1899077,-7.08261227)
    };
    \addlegendentry{$\sigma = 0.2$}

    \addplot[
    color=green,
    mark=o,
    ]
    coordinates {
   (1.51, 1.165) (1.165, 0.4194108709) ( 0.419410870,-0.9566) (-0.956662787,-3.2899077) ( -3.2899077,-7.15261227)
    };
    \addlegendentry{$\sigma = 0.25$}

\end{axis}
\end{tikzpicture}}
\subfigure[]{
\begin{tikzpicture}
\begin{axis}[legend style={font=\tiny},
    xlabel={Iteration count $t$},
    ylabel={$ \max_j \{ \|\theta^{(t+1)}_j - \theta^*_j \| \}$ },
    xmin=0, xmax=50,
    ymin=-0.2, ymax=1.3,
    legend pos=north east,
    ymajorgrids=true,
    xmajorgrids=true,
    grid style=dashed,
]
 
\addplot[
    color=red,
    mark=o,
    ]
    coordinates {
  (1,1.02162667e+00)(2, 7.64912857e-01)(3,5.81823094e-01)
         (4,4.44582634e-01)  (5, 3.40060288e-01) (6,2.75277331e-01)
         (7,2.32005284e-01)(8, 1.96016642e-01)(9,1.68867647e-01)
         (10,1.45531639e-01)(11,   1.25555900e-01)(12,   1.08828061e-01)(13,
         9.45896536e-02)(14,   8.24040419e-02)(15,   7.20073243e-02)(16,6.27099318e-02)(17,   5.49098239e-02)(18,   4.82020701e-02)(19,
         4.23990251e-02)(20,   3.72528460e-02)(21,   3.27623294e-02)(22,
         2.87675316e-02)(23,   2.52780733e-02)(24,   2.22224741e-02)
         (25,
         1.92764028e-02)(26,   1.67669520e-02)(27,   1.45709003e-02)(28,
         1.26754765e-02)(29,   1.10403555e-02)(30,   9.60345571e-03)(31,
         8.36442738e-03)(32,   7.29140677e-03)(33,   6.36035684e-03) (34,
         5.55093547e-03)(35,   4.84735399e-03)(36,   4.16548064e-03)(37,
         3.59317135e-03)(38,   3.10352644e-03)(39,   2.68309579e-03)(40,
         2.32141564e-03)(41,   2.00983541e-03)(42,   1.74109963e-03)(43,
         1.50908292e-03)(44,   1.30859170e-03)(45,   1.13520800e-03)(46,
         9.85163581e-04)(47,   8.55237138e-04)(48,   7.42669567e-04)

    };
    \addlegendentry{$d=50$}

\addplot[
    color= blue,
    mark=diamond,
    ]
    coordinates {
    
    (1,1.16162667e+00)(2,   1.04163821e+00)   (3,9.30127831e-01)(
         4,7.45390389e-01)(   5,5.92415632e-01)(   6,4.58241343e-01)(
         7,2.94074525e-01)(   8,2.16425050e-01)(   9,1.78867647e-01)(10,1.645531639e-01)(   11,1.425555900e-01)( 12,1.18828061e-01)(
         13,8.945896536e-02)(  14,8.64040419e-02)(  15, 7.60073243e-02)(
         16,6.77099318e-02)(  17, 5.89098239e-02)(18,   4.982020701e-02) (19,   3.72528460e-02)(20,   3.27623294e-02)(21,
         2.87675316e-02)(22,   2.52780733e-02)(23,   2.22224741e-02)(24,
         1.92764028e-02)(25,   1.67669520e-02)(26,   1.45709003e-02)(27,
         1.26754765e-02)(28,   1.10403555e-02)(29,   9.60345571e-03)(30,
         8.36442738e-03)(31,   7.29140677e-03)(32,   6.36035684e-03)(33,
         5.55093547e-03)(34,   4.84735399e-03)(35,   4.16548064e-03)(36,
         3.59317135e-03)(37,   3.10352644e-03)(38,   2.68309579e-03)(39,
         2.32141564e-03)(40,   2.00983541e-03)(41,   1.74109963e-03)(42,
         1.50908292e-03)(43,   1.30859170e-03)(44,   1.10520800e-03) (45, 1.10520800e-03)
          (46,
         9.85163581e-04)(47,   8.55237138e-04)(48,   7.42669567e-04)
    };
    \addlegendentry{$d=100$}

    \addplot[
    color=green,
    mark=o,
    ]
    coordinates {
    
    (1,1.25000989e+00)(2,
         1.12152752e+00)(3,   1.00518878e+00)(4,   8.98365520e-01)
         (5,8.03673444e-01)(6,   7.16927909e-01)(7,   6.30996314e-01)(8,
         5.58622493e-01)(9,   4.89647277e-01)(10,   4.31531436e-01)(11,
         3.80223898e-01)(12,   3.35677898e-01)(13,   2.97431811e-01)(14,
         2.64296153e-01)(15,   2.35220954e-01)(16,   2.09390196e-01)(17,
         1.86042522e-01)(18,   1.65679340e-01)(19,   1.47838392e-01)(20,
         1.32077879e-01)(21,   1.17926994e-01)(22,   1.05512167e-01)(23,
         9.45166489e-02)(24,   8.47169500e-02)(25,   7.57425523e-02)(26,
         6.79065493e-02)(27,   6.08966959e-02)(28,   5.46986103e-02)(29,
         4.92145020e-02)(30,   4.43448534e-02)(31,  3.99635140e-02)(32,   3.59913151e-02)(33,
         3.23715203e-02)(34,  2.91713534e-02)(35,   2.63236704e-02)(36,
         2.37823639e-02)(37,   2.15096516e-02)(38,   1.94735221e-02)(39,
         1.76464768e-02)(40,   1.60047101e-02)(41,   1.45275050e-02)(42,
         1.31967603e-02)(43,   1.08957966e-02)(44,   9.02438814e-03)
         
      (45, 7.48808905e-03)
          (46,
         6.22692313e-03)(47,   5.68250591e-03)(48,   5.18800902e-03)

    };
    \addlegendentry{$d=250$}

\end{axis}
\end{tikzpicture}}
\subfigure[]{
\begin{tikzpicture}
\begin{axis}[legend style={font=\tiny},
    xlabel={$\log ( \max_j \{ \|\theta^{(t)}_j - \theta^*_j \| \} )$},
    ylabel={$\log ( \max_j \{ \|\theta^{(t+1)}_j - \theta^*_j \| \} )$ },
    xmin=-3.4, xmax=0.7,
    ymin=-3.2, ymax=0.6,
    legend pos=north west,
    ymajorgrids=true,
    xmajorgrids=true,
    grid style=dashed,
]
 
\addplot[
    color=red,
    mark=o,
    ]
    coordinates {

  (0.34657359,  0.14278554)(0.14278554, -0.06224563)(-0.06224563, -0.24996209)(-0.24996209, -0.41189006)(-0.41189006,
       -0.56667079)(-0.56667079, -0.73899015)(-0.73899015, -0.88769828)(-0.88769828, -1.04723158)(-1.04723158, -1.20001584)( -1.20001584,
       -1.35632364)(-1.35632364, -1.50352481)(-1.50352481, -1.65728063)(-1.65728063, -1.80197766)(-1.80197766, -1.93507441)(-1.93507441,
       -2.05976912)(-2.05976912, -2.1792287)(-2.1792287 , -2.29764125)(-2.29764125, -2.42380908)(-2.42380908, -2.54690996)(-2.54690996,
       -2.66800827)(-2.66800827, -2.7874863)(-2.7874863 , -2.92346283)(-2.92346283, -3.06099054)(-3.06099054, -3.19738121)

    };
    \addlegendentry{$d=50$}

\addplot[
    color= blue,
    mark=diamond,
    ]
    coordinates {
    (6.93147181e-01,   3.90209780e-01)(3.90209780e-01,   1.73424199e-01)(1.73424199e-01,  -1.7510556e-01)(-1.7510556e-01,  -3.90300333e-01)(-3.90300333e-01,
        -5.42720773e-01)(-5.42720773e-01,  -6.90503681e-01)(-6.90503681e-01,  -8.37669171e-01)(-8.37669171e-01,
        -9.79724699e-01)( -9.79724699e-01,  -1.11314673e+00)(-1.11314673e+00,  -1.23751918e+00)(-1.23751918e+00,
        -1.36058964e+00)(-1.36058964e+00,  -1.48852036e+00)(-1.48852036e+00,  -1.61386355e+00)(-1.61386355e+00,
        -1.74199555e+00)(-1.74199555e+00,  -1.87181284e+00)(-1.87181284e+00,  -1.99648300e+00)(-1.99648300e+00,
        -2.12081252e+00)(-2.12081252e+00,  -2.24229654e+00)(-2.24229654e+00,  -2.36353872e+00)(-2.36353872e+00,
        -2.49521359e+00)(-2.49521359e+00,  -2.62501938e+00)(-2.62501938e+00,  -2.75960253e+00)(-2.75960253e+00,
        -2.89040955e+00)(-2.89040955e+00,  -3.02104933e+00)
    };
    \addlegendentry{$d=100$}

    \addplot[
    color=green,
    mark=o,
    ]
    coordinates {
   (0.69179179,  0.58458761)(0.58458761,
        0.48819941)(0.48819941,  0.39855741)(0.39855741,  0.30824143)(0.30824143,  0.21243723)(0.21243723,  0.11755429)(0.11755429,
        0.02503039)(0.02503039, -0.06644474)(-0.06644474, -0.16123063)(-0.16123063, -0.25831131)(-0.25831131, -0.3580291)(-0.3580291 ,
       -0.46048038)(-0.46048038, -0.55709383)(-0.55709383, -0.65461168)(-0.65461168, -0.75102489)(-0.75102489, -0.84396401)(-0.84396401,
       -0.93729249)(-0.93729249, -1.02728466)(-1.02728466, -1.12467717)(-1.12467717, -1.22193158)(-1.22193158, -1.3170877)(-1.3170877 ,
       -1.41237518)(-1.41237518, -1.50604708)(-1.50604708, -1.60010283)(-1.60010283, -1.69229454)(-1.69229454, -1.78408158)(-1.78408158,
       -1.87790426)(-1.87790426, -1.97032467)(-1.97032467, -2.06308162)(-2.06308162, -2.15624736)( -2.15624736, -2.24806468)(-2.24806468,
       -2.33885897)(-2.33885897, -2.42950567)(-2.42950567, -2.51975678)(-2.51975678, -2.61571586)(-2.61571586, -2.71065425)(-2.71065425,
       -2.80696082)(-2.80696082, -2.9030375)(-2.9030375 , -2.99948999)(-2.99948999, -3.09644304)(-3.09644304, -3.19185196)
    };
    \addlegendentry{$d=250$}
    \end{axis}
\end{tikzpicture}}
\caption{Noisy AM and comparison with gradient based heuristic---in panel (a), we plot the (log) optimization error $\log ( \max_j \{ \|\theta^{(t)}_j - \theta^{(T)}_j \| \} )$ and show that AM retain the super-linear speed of convergence even in the presence of noise. Here we fix $d = 250$ and $n = 6 \, d$ and vary over $\sigma$ $(0.1,0.2,0.25)$. In Panel (b) we show that gradient based heuristic (Algorithm~\ref{alg:heu} indeed recovers the true parameters. We fix $n = 6 \, d$ and vary $d (=50,100,200)$. We also tune the step-size to fasten Algorithm~\ref{alg:heu}. In Panel (c) we demonstrate that the rate of convergence of Algorithm~\ref{alg:heu} is truly linear. We plot $\log ( \max_j \{ \|\theta^{(t+1)}_j - \theta^*_j \| \} )$ with respect to $\log ( \max_j \{ \|\theta^{(t)}_j - \theta^*_j \| \} )$ and the slope of the line is close to $1$ implying linear rate of convergence.}
\label{fig:gd_conv}
\end{figure*}

\paragraph{Convergence of AM for $2$ mixture:} We first show the convergence of Algorithm~\ref{alg:am} for a mixture of $2$ linear regression in the noiseless setting. The results are shown in Figure~\ref{fig:am_conv}.  In panel (a) of Figure~\ref{fig:am_conv}, we plot $\max\{\| \theta^{(t)}_1 - \theta^*_1 \|,\| \theta^{(t)}_1 - \theta^*_2 \| \}$ with respect to iterations of Algorithm~\ref{alg:am}, where $ \{\theta^{(t)}_1, \theta^{(t)}_2 \}$ is the output at the $t$-th iterate of the algorithm. We consider different values of $d$ (specifically $50, 100, 250 $ and $500$) and choose the sample size $n = 6\,d$. We observe that the iterates converge to $0$ very quickly, and hence Algorithm~\ref{alg:am} guarantees perfect recovery of $\{ \theta^*_1,\theta^*_2 \}$. In fact, we see that AM takes at most $6$ steps to converge.

\paragraph{Super linear convergence for $2$ mixture:}
In Figure~\ref{alg:am} (b), we characterize the rate of convergence of Algorithm~\ref{alg:am}. Here, we plot $\log \left( \max\{\| \theta^{(t+1)}_1 - \theta^*_1 \|,\| \theta^{(t+1)}_2 - \theta^*_2 \| \} \right)$ with respect to $\log \left( \max\{\| \theta^{(t)}_1 - \theta^*_1 \|,\| \theta^{(t)}_2 - \theta^*_2 \| \} \right)$. Note that the slope of this plot quantifies the rate of convergence of the underlying iterative algorithm that produces the iterates $\{\theta^{(j)}_1, \theta^{(j)}_2 \}$ for $j = 0,1,\ldots $. Algorithms with linear rate of convergence will result in slope $1$. Any slope strictly greater than $1$ implies super-linear convergence.

We set $d = 250, 500, 1000$ and $2000$ and take $n = 6 \, d$. The results are shown in Figure~\ref{fig:am_conv} (b). We observe that the slope of the line is around $1.7 - 1.8$, which validates our theory of super-linear convergence. In Theorem~\ref{thm:sup_lin} and \ref{thm:quadratic}, we prove that the exponent of convergence is $1.5$ and $2$ under different settings. However, in the simulations, we observe the exponent of (super-linear) convergence is a constant between $1.5$ and $2$.

\paragraph{Super linear convergence for more than $2$ mixtures:}
We now show empirically that our theory of super-linear convergence holds for mixture of more than $2$ linear regression. For this setting we use an extension of Algorithm~\ref{alg:am} tailored to more than $2$ mixtures. This is precisely Algorithm 3 of Yi et al (2016) (\cite{yimany}). In Figure~\ref{fig:am_conv} (c), we consider a mixture of $3$ linear regression. We consider $d = 200,250$ and $500$ and take $n = 15 \, d$. Similar to the $2$ mixture setting, we are interested in the rate of convergence and hence we plot $\log \left( \max_{j \in \{1,2,3\}} \{ \|\theta_j^{(t+1)} - \theta^*_j \| \} \right)$ with respect to $\log \left( \max_{j  \in \{1,2,3\}} \{ \|\theta_j^{(t)} - \theta^*_j \| \} \right)$. From Figure~\ref{fig:am_conv} (c), we see that the plot is linear with slope around $1.7 - 1.85$. Hence, AM retains the super linear speed of convergence for an arbitrary number of mixtures of linear regression.

\paragraph{Algorithm~\ref{alg:am} with noisy observations:}
We now consider the setting where $w_i \neq 0$ in equation~\eqref{eqn:model}. In particular we assume $w_i \stackrel{i.i.d}{\sim} \NORMAL(0,\sigma^2)$. Under this setting, we can only recover  $\{ \theta^*_1, \theta^*_2 \}$ upto an error floor depending on $(\sigma, n, d)$. To demonstrate super-linear convergence in this setting, we first estimate the error floor by letting Algorithm~\ref{alg:am} run for $T = 50$ iterations. We now construct the \emph{optimization-error} $\max_{j \in \{1,2\}} \{ \|\theta_j^{(t)} - \theta^{(T)}_j \| \}$. In Figure~\ref{fig:gd_conv} (a), we plot the variation of the (logarithm of) \emph{optimization-error} in the $t+1$-th iteration with respect to $t$-th iteration. We notice that the dependence is linear (for different values of $\sigma$ ($=0.1,0.2,0.25$)) with a slope of $1.8 - 1.85$. This observation validates the fact that Algorithm~\ref{alg:am} retains the super-linear convergence in the presence of noise.

\paragraph{Comparison with gradient based heuristic:}
Finally, we compare Algorithm~\ref{alg:am} with a gradient based heuristic. The heuristic algorithm  is described in Algorithm~\ref{alg:heu}.

\begin{algorithm}[t]
  \caption{Gradient based heuristic for mixed linear regression}
  \begin{algorithmic}[1]
  \INPUT Covariate response pairs $(\meas_i,y_i)_{i=1}^n$, initialization $(\widehat{\theta}^{(0)}_1,\widehat{\theta}^{(0)}_2)$, step size $\gamma$, number of rounds $T$
  
   \STATE \textbf{for} $t = 0,\ldots, T-1$ \textbf{do} 
   \STATE \hspace{ 4mm} $z_i^{(t)} = \mathrm{argmin}_{j \in \{0,1\}} | y_i - \langle x_i, \widehat{\theta}^{(t)}_j \rangle|;\, i \in [n]$
   \STATE \hspace{4mm} $\widehat{\theta}^{(t+1)}_j = \widehat{\theta}^{(t)}_j - \gamma \nabla \bigg (  (y_i - \langle x_i,\theta \rangle )^2 \ind{z_i^{(t)} = j} \bigg );$ \\
   \hspace{20 mm} for $j \in \{0,1\}$
   
   \STATE \textbf{end for}
 
  \OUTPUT $(\widehat{\theta}^{(T)}_1,\widehat{\theta}^{(T)}_2)$
  \end{algorithmic}\label{alg:heu}
\end{algorithm}

In Algorithm~\ref{alg:heu}, we simply replace the least-squares step by a gradient step. We study this algorithm empirically in the noiseless setting for a mixture of $2$ linear regressions. In Figure~\ref{fig:gd_conv} (b), we show that Algorithm~\ref{alg:heu} indeed converges. We set $d =50,100,250$ and take $n =6 \,d$. Furthermore, we increase the step size $\gamma$ until the algorithm starts oscillating and choose the largest stepsize for which Algorithm~\ref{alg:heu} converges. We observe that even with the step-size tuning, Algorithm~\ref{alg:heu} takes a lot of iterations compared to Algorithm~\ref{alg:am} (Figure~\ref{fig:am_conv} (a)) in identical settings.

We now characterize the rate of convergence of Algorithm~\ref{alg:heu}. In order to do so, similar to the previous scenarios, we plot $\log \left( \max_{j  \in 1,2} \{ \|\theta_j^{(t+1)} - \theta^*_j \| \} \right)$ with respect to $\log \left( \max_{j  \in 1,2} \{ \|\theta_j^{(t)} - \theta^*_j \| \} \right)$. This is shown in Figure~\ref{fig:gd_conv} (c). We observe that the dependence is linear with slope roughly equal to $1$. This shows that the rate of convergence of the gradient based heuristic is truly linear.

Finally, we compare the iteration complexity and wall-clock time of AM (Algorithm~\ref{alg:am}) with the gradient based heuristic (Algorithm~\ref{alg:heu}).  The results are tabulated in Table~\ref{table:time_compare}. Here, we fix a target precision of $0.001$. We observe that compared to AM, Algorithm~\ref{alg:heu} takes $9 \times$ iterations to recover the true parameter within the given precision. We also compare the algorithms in terms of wall-clock time. We found that on average gradient based heuristic takes $6 \times$ time compared to AM, even when we tune and choose the largest step-size. These results imply that AM is a much faster algorithm compared to the gradient based heuristic in terms of both number of iterations and total wall-clock time.

\begin{table}[t!]
\caption{Comparison between AM (Algorithm~\ref{alg:am}) and GD based heuristic (Algorithm~\ref{alg:heu}) in terms of iteration complexity and wall-clock time.} \label{table:time_compare}
\begin{center}
\begin{tabular}{|c |c |c |c | }
\hline
\textbf{DIMEN-} &\textbf{ALGO-}  &\textbf{ITERA-} &\textbf{WALL-CLO} \\
 \textbf{SION ($d$)}  & \textbf{RITHM} & \textbf{TION} & \textbf{CK (SEC)} \\
\hline
$50$  & AM  &$5$     & $ 0.015$ \\
$50$ & GD  & $45$    &  $0.125$    \\
\hline
$100$  & AM  &$5$     & $ 0.094$ \\
$100$ & GD  & $47$    &  $0.659$    \\
\hline
$250$  & AM  &$6$     & $ 0.953$ \\
$250$ & GD  & $48$    &  $6.546$    \\
\hline
\end{tabular}
\end{center}
\end{table}

 \section{Proof of Theorem~\ref{thm:sup_lin}}
 \label{sec:proof_thm_one}
 
  We now prove Theorem~\ref{thm:sup_lin}. To simplify notation, we use the shorthand $\mathsf{dist}:=\mathsf{dist}(\theta_1,\theta_2)$ and $\mathsf{dist}^+ := \mathsf{dist}(\theta_1^+,\theta_2^+)$. Also, we drop the superscript in $\meas_i^{(t)}$ and $y_i^{(t)}$. Recall that the sample complexity for this iteration is $n/T = n_1$.

  Here we retain the notation of \cite{yitwo}. To that end, let us denote the set of indices $J_1^*$ and $J_2^*$ corresponding to observations coming from $\sigstar_1$ and $\sigstar_2$ respectively. Similarly, we define $J_1$ and $J_2$ corresponding to the iterate of AM for $\theta_1$ and $\theta_2$ respectively. Hence, we have
 \begin{align*}
 J_1^* = \{ i \in [n_1]: y_i = \langle \meas_i, \sigstar_1 \rangle \}
 \end{align*}
and similarly, using the criteria of Algorithm~\ref{alg:am}, we have
\begin{align*}
J_1 = \{ i \in [n_1]: (y_i - \langle \meas_i, \theta_1 \rangle)^2 < (y_i - \langle \meas_i,\theta_2 \rangle)^2 \}.
\end{align*} 
 We can define $J^*_2$ and $J_2$ in a similar way. We also define a diagonal matrix $W \in \real^{n_1 \times n_1}$ such that $W_{ii} = 1$ if $i \in J_1$ and $0$ if $i \in J_2$. Similarly, we define $W^*$ such that $W^*_{ii} = 1$ if $i \in J^*_1$ and $0$ otherwise. With this new notation, it immediately follows that $\theta^+_1$ is the least squares solution to $Wy = W X \theta$, and hence 
 \begin{align*}
 \theta^+_1 = (X^\top W X)^{-1} X^\top W y
 \end{align*}
where the $n_1$ dimensional vector $y = [y_1 \ldots y_{n_1}]^\top$ and we use the fact that $W^2 =W$. Similarly we observe that $\theta^+_2$ is the least squares solution of  $(I-W)y = (I-W)X\theta$.

With this, the observation vector $y$ can be written as
\begin{align*}
y = W^* X \theta^*_1 + (I-W^*)X \theta^*_2,
\end{align*}
and hence substituting $y$ in the closed form expression for $\theta^+_1$, we obtain 
\begin{align*}
\theta^+_1 - \theta^*_1 = (X^\top W X)^{-1} X^\top (WW^* - W)X(\theta^*_1- \theta^*_2). 
\end{align*}
Let $\mathcal{S} = J_1 \cap J^*_2$. We have
\begin{align}
& \| X(\theta^+_1 - \theta^*_1)\|^2 \nonumber \\
& = \| X (X^\top W X)^{-1} X^\top (WW^* - W)X(\theta^*_1- \theta^*_2) \|^2 \label{eqn:substitution} \\
& = \| X_{J_1} X_{J_1}^\dagger (WW^* - W)X(\theta^*_1- \theta^*_2) \|^2 \label{eqn:non-zero-val} \\
& \leq \|X_\mathcal{S}(\theta^*_1 - \theta^*_2) \|^2 \label{eqn:proj_non_expan} \\
& = \sum_{i \in \mathcal{S}} \langle \meas_i, \theta^*_1 - \theta^*_2 \rangle^2 \nonumber
\end{align} 
where equation~\eqref{eqn:substitution} follows from substituting $\theta^+_1 - \theta^*_1$, equation~\eqref{eqn:non-zero-val} follows from the definition of $J_1$. Equation~\eqref{eqn:proj_non_expan} follows from the facts that equation~\ref{eqn:non-zero-val} is non-zero only when $i \in \mathcal{S}$ and $X_{J_1} X_{J_1}^\dagger$ is a projection matrix and hence non-expansive. Continuing, we obtain
\begin{align}
\| X(\theta^+_1 - \theta^*_1)\|^2 & \leq \sum_{i \in \mathcal{S}} \langle \meas_i, \theta^*_1 - \theta^*_2 \rangle^2 \nonumber \\
& = \sum_{i \in \mathcal{S}} \langle \meas_i, \theta^*_1 - \theta_1 + \theta_1 - \theta^*_2 \rangle^2 \nonumber \\
& \leq \sum_{i \in \mathcal{S}} \bigg ( 2 \langle \meas_i, \theta^*_1 - \theta_1 \rangle^2 + 2 \langle \meas_i, \theta^*_2 - \theta_1 \rangle^2 \bigg ) \nonumber
\end{align}  
 Now, recall that if $i \in \mathcal{S}$, $y_i = \langle x_i, \theta^*_2 \rangle$. Also, we have
 \begin{align*}
 & (y_i - \langle \meas_i, \theta_1 \rangle)^2 < (y_i - \langle \meas_i,\theta_2 \rangle)^2 \nonumber \\
 & \Rightarrow \langle \meas_i, \theta^*_2 - \theta_1 \rangle^2 < \langle \meas_i, \theta^*_2 - \theta_2 \rangle^2.
 \end{align*}
 Substituting the above, we have
 \begin{align}
 \| X(\theta^+_1 - \theta^*_1)\|^2 & \leq 2 \sum_{i \in \mathcal{S}}  \langle \meas_i, \theta^*_1 - \theta_1 \rangle^2 \nonumber \\
 & + 2 \sum_{i \in \mathcal{S}} \langle \meas_i, \theta^*_2 - \theta_2 \rangle^2 \label{eqn:pred_err}
 \end{align}
We now concentrate on the first term of the right hand side of equation~\eqref{eqn:pred_err}. Recall that from the definition of $\mathsf{dist}$, we have $\|\theta_1 - \theta^*_1 \|^2 \leq \mathsf{dist}^2$. Hence, we obtain
\begin{align*}
\sum_{i \in \mathcal{S}}  \langle \meas_i, \theta^*_1 - \theta_1 \rangle^2 & \leq (\mathsf{dist}^2)  \left(\sum_{i \in \mathcal{S}} \langle \meas_i, \frac{\theta_1 -\theta^*}{\|\theta_1 -\theta^*\|} \rangle^2 \right)
\end{align*}

Let us define the unit vector $u = \frac{\theta_1 -\theta^*}{\|\theta_1 -\theta^*\|}$. Since we re-sample at each iteration of the AM algorithm, $x_i$ is independent of $u$. Furthermore, conditioned on the fact that $i \in \mathcal{S}$, the distribution of $\meas_i$ is no longer Gaussian. To this end, \cite[Lemma 15(b)]{yimany} shows that the distribution of $x_i$ is $c$-sub-Gaussian, implying that $\langle x_i, u \rangle$ is a $c$ sub-Gaussian random variable (here $c$ is a constant) with $\E [\langle x_i, u \rangle^2] \leq C$, where $C$ is a constant. Moreover, using \cite[Lemma 2.7.6]{vershynin2018high}, the distribution of $\langle x_i, u \rangle^2$ is $(\tilde{c_1},\tilde{c_2})$ sub-exponential, where $\tilde{c_1}$ and $\tilde{c_2}$ are constants. Similar argument holds for the second term in the right hand side of equation~\eqref{eqn:pred_err}.

We now use the following Lemma which gives a high probability upper-bound on $|\mathcal{S}|$.
  \begin{lemma}
  \label{lem:error_set}
We have
\begin{align*}
|\mathcal{S}| \leq C_1 \frac{ p_2 n_1 \,\, \mathsf{dist}}{\|\theta^*_1 - \theta^*_2 \|}
\end{align*} 
with probability exceeding $1- n_1^{-10}$ provided $\mathsf{dist} \geq c \max\{p_1,p_2\} \sqrt{\frac{\log n_1}{n_1}}$.
  \end{lemma}
We first take this lemma for granted and conclude the proof of the theorem. Let $k_0:= C_1 \frac{ p_2 \mathsf{dist}\, n_1}{\|\theta^*_1 - \theta^*_2 \|}$. We have
 \begin{align*}
\Prob \left( \sum_{i=1}^{|\mathcal{S}|}\langle x_i, u \rangle^2 \geq t \right) \leq \Prob ( \sum_{i=1}^{k_0}\langle x_i, u \rangle^2 \geq t ) + \Prob(|\mathcal{S}| \geq k_0)
 \end{align*}
 Provided $\mathsf{dist} \geq c \max \{p_1,p_2 \} \sqrt{\frac{\log n_1}{n_1}}$ and choosing $t= \frac{3 k_0 \log n_1}{2}$, sub-exponential concentration (\cite[Chapter 2]{wainwright2019high}) along with Lemma~\ref{lem:error_set} yields
 \begin{align*}
 \Prob ( \sum_{i=1}^{|\mathcal{S}|}\langle x_i, u \rangle^2 \geq t ) &\leq \Prob ( \sum_{i=1}^{k_0}\langle x_i, u \rangle^2 \geq \frac{3k_0 \log n_1}{2} ) + c_1 n_1^{-10} \nonumber \\
 & \leq c n_1^{-10}  + c_1 n_1^{-10}.
 \end{align*}
Similar expression holds for the second term of equation~\eqref{eqn:pred_err}. Substituting this in equation~\eqref{eqn:pred_err}, we obtain
 \begin{align}
 \| X(\theta^+_1 - \theta^*_1)\|^2 \leq 6 C_1 (\mathsf{dist})^3 \frac{p_2 n_1 \log n_1}{\|\theta^*_1 - \theta^*_2 \|}. \label{eqn:final_pred_err}
 \end{align}
We now convert this prediction error to  estimation error via exploiting the spectral properties of the Gaussian random matrix $X$. We have
 \begin{align*}
 \| X(\theta^+_1 - \theta^*_1)\|^2 \geq \lambda_{\min}(X^\top X) \|\theta^+_1 - \theta^*_1\|^2.
 \end{align*}
Since $X \in \real^{n \times d}$ is a Gaussian random matrix, using \cite{vershynin2010introduction}, the minimum singular value is
\begin{align*}
\sigma_{\min}(X) \geq \sqrt{n_1} - \sqrt{d} - t
\end{align*} 
with probability exceeding $1-\exp\{ - c t^2\}$.

Using the fact that $n_1 \geq C d$, and substituting $t = \sqrt{n_1}/\sqrt{C}$, we obtain
\begin{align*}
\lambda_{\min}(X^\top X) \geq c \, n_1
\end{align*} 
 with probability exceeding $1- \exp\{- c_1 \, n_1\}$. Putting everything together, we have
 \begin{align*}
 \|\theta^+_1 - \theta^*_1\|^2 \leq  \frac{6 C_1 p_2 \log n_1}{c \|\theta^*_1 - \theta^*_2 \|} \mathsf{dist}^3 \leq \frac{1}{4} \frac{\log n_1}{\|\theta^*_1 - \theta^*_2\|} \mathsf{dist}^3 
 \end{align*}
 where we use the fact that $p_2 \leq 1$, and choose appropriate constants $c$ and $C_1$.
 
 Similarly we prove an upper bound for $\|\theta^+_2 - \theta^*_2\|^2$, and hence the theorem follows.

\section{CONCLUSION}
We prove the super linear convergence of AM for noiseless mixture of $2$ linear regressions. However, in experiments, we see that the super linear rate retains for noisy setting and even for more than $2$ mixtures. Providing theoretical guarantees in these settings will be our immediate future works. In experiments we also observe that sample-split is unnecessary and the exponent of convergence is around $1.75-1.8$. Is the exponent really $1.5$, or a finer analysis can improve this? We leave this questions as our future endeavors.

\section{ACKNOWLEDGMENTS} We thank Ashwin Pananjady and Aditya Guntuboyina for their insightful comments. The idea of super-linear convergence of AM is originated with Ashwin Pananjady for the problem of real phase retrieval. We also thank the reviewers of AISTATS 2020 for their insightful remarks.

\bibliographystyle{alpha}
\bibliography{mix-lin}

\clearpage
\onecolumn

 
  \subsection{Proof of Lemma~\ref{lem:error_set}:}
 We first upper bound the quantity $\E |\mathcal{S}|$. Then invoking Hoeffding's inequality, we obtain a high probability bound on $|\mathcal{S}|$. We have
 \begin{align*}
 \E |\mathcal{S}| & = \E \sum_{i=1}^{n_1} \ind{i \in J_1 \cap J^*_2} \nonumber \\
 & = \sum_{i=1}^{n_1} \Prob( i \in J^*_2 ) \Prob (i \in J_1 | i \in J^*_2 )
 \end{align*}
The event $i \in J_1 $, conditioned on the event $i \in J^*_2$ is equivalent to the event 
\begin{align*}
\langle \meas_i,\theta_1 - \theta_2^* \rangle^2 < \langle \meas_i, \theta_2 -\theta^*_2 \rangle^2.
\end{align*}
 Hence, we have
\begin{align*}
\E (\ind{ i \in J_1 \cap J^*_2 }) = & \Prob( i \in J^*_2 ) \times \nonumber \\
& \Prob \left( \langle \meas_i,\theta_1 - \theta_2^* \rangle^2 < \langle \meas_i, \theta_2 -\theta^*_2 \rangle^2 \right).
\end{align*}
Now, from the initialization condition, we have
\begin{align*}
\mathsf{dist} \leq \frac{\| \theta^*_1 - \theta^*_2 \|}{2},
\end{align*} 
 and hence $\| \theta_1 - \theta^*_2 \| > \|\theta_2 -\theta^*_2 \|$. Using this fact and invoking Lemma 1 of \cite{yitwo} yields
 \begin{align*}
 \Prob \left( \langle \meas_i,\theta_1 - \theta_2^* \rangle^2 < \langle \meas_i, \theta_2 -\theta^*_2 \rangle^2 \right) \leq \frac{\|\theta_2 -\theta^*_2 \|}{\| \theta_1 - \theta^*_2 \|}.
 \end{align*}
Also, we have
\begin{align*}
\| \theta_1 - \theta^*_2 \| & = \|\theta_1 - \theta^*_1 + \theta^*_1 - \theta^*_2 \| \nonumber \\
& \geq \|\theta^*_1 - \theta^*_2 \| -  \|\theta_1 - \theta^*_1\| \geq \frac{1}{2} \|\theta^*_1 - \theta^*_2 \|
\end{align*} 
 where the last inequality follows from the initialization condition. Substituting this, we obtain
 \begin{align*}
 \Prob \left( \langle \meas_i,\theta_1 - \theta_2^* \rangle^2 < \langle \meas_i, \theta_2 -\theta^*_2 \rangle^2 \right) & \leq \frac{2 \|\theta_2 -\theta^*_2 \|}{\|\theta^*_1 - \theta^*_2 \|} \nonumber \\
 & \leq \frac{2 \,\,\, \mathsf{dist}}{\|\theta^*_1 - \theta^*_2 \|}.
 \end{align*}
 So, we finally have
 \begin{align*}
 \E |\mathcal{S}| \leq \frac{2 p_2 n_1 \,\, \mathsf{dist}}{\|\theta^*_1 - \theta^*_2 \|}.
 \end{align*}
 We now invoke the Hoeffding's inequality to obtain
 \begin{align*}
 \Prob \bigg ( ||\mathcal{S}| - \E |\mathcal{S}|| > n_1 t \bigg ) \leq 2 \exp\{- c n_1 t^2 \}
 \end{align*}
Substituting $t = C \frac{ p_2 \, \mathsf{dist}}{\|\theta^*_1 - \theta^*_2 \|}$, we obtain
\begin{align*}
|\mathcal{S}| \leq C_1 \frac{ p_2 n_1 \,\, \mathsf{dist}}{\|\theta^*_1 - \theta^*_2 \|}
\end{align*} 
with probability exceeding $1- c_1 n_1^{-10}$ as long as $\mathsf{dist} \geq C \max\{p_1,p_2 \} \sqrt{\frac{\log n_1}{n_1}}$.

 \subsection{Proof of Theorem~\ref{thm:quadratic}}
We follow the same setup and notations as in the proof of Theorem~\ref{thm:sup_lin}. This proof borrows a few technical details from the proof of \cite[Theorem 1]{yitwo}. Recall that we have
\begin{align*}
\theta^+_1 - \theta^*_1 = (X^\top W X)^{-1} X^\top (WW^* - W)X(\theta^*_1- \theta^*_2). 
\end{align*}
Hence, we obtain $\| \theta^+_1 - \theta^*_1 \| \leq A B $, where 
\begin{align*}
& A = \| (X^\top W X)^{-1} \| \quad \text{and,} \\
& B = \|X^\top(WW^* -W)X(\theta^*_1-\theta^*_2) \|.
\end{align*}
\paragraph*{Bounding $A$:} We just use the bound in \cite[Theorem 1]{yitwo} to control $A$, which yields the following. If $n_1 \geq C d$,  we have 
\begin{align*}
A \leq \frac{C}{n_1}
\end{align*}
with probability exceeding $1-c\exp\{-c_1 d \}$.
\paragraph*{Bounding $B$:} Let
\begin{align*}
Q = X^\top (WW^* -W)X.
\end{align*}
Using the proof steps of \cite[Theorem 1]{yitwo}, we obtain
\begin{align*}
B \leq 2 \sigma_{\max}(Q)\, \mathsf{dist}.
\end{align*}
Note that the $Q$ is non-zero when $i \in J_1 \cap J^*_2$. Using Lemma 1 of \cite{yitwo} in conjunction with \cite{vershynin2010introduction} yields
\begin{align*}
\sigma_{\max}(Q) \leq c \max\left(d,|J_1 \cap J^*_2|\right).
\end{align*}
In the proof of Lemma~\ref{lem:error_set}, we observe that
 \begin{align*}
 \E |J_1 \cap J_2^*| \leq  \frac{2 p_2 n_1 \,\, \mathsf{dist}}{\|\theta^*_1 - \theta^*_2 \|} \leq C n_1 \, \mathsf{dist}.
 \end{align*}
 Hence applying Hoeffding's inequality yields
\begin{align*}
|J_1 \cap J_2^*| \geq C_1 n_1 \, \mathsf{dist}
\end{align*}
with probability exceeding $1- c n_1^{-10}$ provided $\mathsf{dist} \geq C \max\{p_1,p_2\} \sqrt{\frac{\log n_1}{n_1}}$.

Now, additionally if $\mathsf{dist} \geq (\frac{1}{C_1}) \frac{d}{n}$, we obtain
\begin{align*}
|J_1 \cap J_2^*| \geq d.
\end{align*}
 With this, we have
\begin{align*}
B \leq  c|J_1 \cap J^*_2| \, \mathsf{dist} \leq c_1 n \, \mathsf{dist}^2.
\end{align*}
where we use Lemma~\ref{lem:error_set} once again in the last step. Putting everything together, we obtain
\begin{align*}
\| \theta^+_1 - \theta^*_1 \| \leq \frac{C}{ n_1} c_1 n_1 \, \mathsf{dist}^2 \leq \frac{1}{2} \, \mathsf{dist}^2,
\end{align*}
provided
\begin{align*}
\mathsf{dist} \geq \max \left \lbrace C \max\{p_1,p_2\} \sqrt{\frac{\log n_1}{n_1}}, \frac{d}{C_1 n_1}\right \rbrace.
\end{align*}
The constants are chosen such that $C \, c_1 \leq 1/2$. Similarly we show an equivalent upper bound on $\| \theta^+_2 - \theta^*_2 \|$, thus yielding the theorem.
 
 Note that, the condition $\mathsf{dist} \geq \frac{d}{C_1 n_1}$ on the $\mathsf{dist}$ function is quite strong since we are interested in the information theoretic minimum sample regime with $n_1 = \mathcal{O}(d)$.

\end{document}

%% file: command-file-arXiv.tex
\usepackage{comment}
\usepackage{epsf}
\usepackage{graphicx}

\usepackage{subfigure}

\usepackage{color}

\usepackage{mathtools}
\usepackage{amsfonts}
\usepackage{amsthm}
\usepackage{amsmath, bm}
\usepackage{amssymb}

\usepackage{textcomp}
\usepackage{xcolor}

\DeclareSymbolFont{extraup}{U}{zavm}{m}{n}
\DeclareMathSymbol{\vardiamond}{\mathalpha}{extraup}{87}

\newtheorem{theorem}{Theorem}

\newtheorem{assumption}{Assumption}

\newtheorem{lemma}{Lemma}

\long\def\comment#1{}

\newcommand{\sigit}[1]{\ensuremath{\theta^{(#1)}}}

\newcommand{\inprod}[2]{\ensuremath{\langle #1 , \, #2 \rangle}}

\newcommand{\E}{\ensuremath{{\mathbb{E}}}}

\newcommand{\Prob}{\ensuremath{{\mathbb{P}}}}

\newcommand{\ind}[1]{\ensuremath{{\mathbf{1}\left\{ #1 \right\}}}}

\newcommand{\NORMAL}{\ensuremath{\mathcal{N}}}

\newcommand{\sigstar}{\ensuremath{\theta^*}}

\newcommand{\meas}{\ensuremath{x}}

\newcommand{\obs}[2]{\ensuremath{ \inprod{\meas_{#1}}{\sigstar_{#2}}}}

\newcommand{\real}{\ensuremath{\mathbb{R}}}

